\documentclass[runningheads]{llncs}
\usepackage[T1]{fontenc}
\usepackage{graphicx}
\usepackage{booktabs}

\usepackage{multirow}

\usepackage{algorithm}
\usepackage{algorithmic}
\usepackage{amssymb}
\usepackage{cite}

\usepackage{wrapfig}

\usepackage[misc]{ifsym}

\usepackage{mwe}
\usepackage{amsmath}

\usepackage{hyperref}

\newcommand{\ie}{\textit{i}.\textit{e}.}

\begin{document}

\title{From Deep to Shallow: Unconstrained and Efficient Layer Merging Strategy}

\titlerunning{From Deep to Shallow: Unconstrained and Efficient Layer Merging Strategy}
% If the full title of your paper is short enough to also fit in the running head, you can omit the abbreviated paper title here. You can check as follows: if you comment out the \titlerunning line, something will appear in the header of all odd-numbered pages of your PDF from page 3 onward. This something is either the full title (in which case all is well), or the error message "Title Suppressed Due to Excessive Length". If this error message appears, you're going to want to provide an abbreviated title within the \titlerunning command, because if you won't do it, Springer will do it for you.

%N.B.: Author information (both in the \author{} and \authorrunning{} command) should only be present in the Camera-Ready Version of your paper. The version that you initially submit for review, ought to be double-blind. So, when initially submitting your paper, use:
%\author{Author information scrubbed for double-blind reviewing}
\author{Petro Shulzhenko\inst{1}\orcidID{0009-0001-6278-5881} \and
Gabriele Spadaro\inst{1,2}\orcidID{0009-0008-4786-1074} \and
Enzo Tartaglione\inst{1}\orcidID{0000-0003-4274-8298}}
\authorrunning{P. Shulzhenko, G. Spadaro, and E. Tartaglione}
% If there are two authors, write 'A.L. Benjamin and C.C. Broadus Jr.'
% If there are more than two authors, '[...] et al.' is used.

\institute{LTCI, Télécom Paris, Institut Polytechnique de Paris, Palaiseau, France \\
\email{\{name.surname\}@telecom-paris.fr}\\ \and
University of Turin, Italy\\
\email{gabriele.spadaro@unito.it}}
\maketitle              % typeset the header of the contribution

\begin{abstract}

Although Deep Neural Networks have become foundational in many areas of Machine Learning, high computational demands limit their application in resource-constrained environments.
To address this issue, depth compression methods have been proposed to identify and linearize redundant activation functions, thereby allowing for the merging of layers without intermediate non-linearities. However, these methods face two key challenges: they cannot be directly applied to convolutions with padding due to the absence of an analytical solution for merging these layers, and they typically increase the kernel size of merged layers, thus limiting speed-up gains.
To overcome these limitations, we propose an efficient strategy that enables merging of layers without an existing analytical solution, and also without increasing kernel size.
We validate our approach across multiple architectures and datasets, and measure inference speed-up gains on real embedded platforms. We publicly released the code at \href{https://github.com/ShulzhenkoPetr/deep-to-shallow}{https://github.com/ShulzhenkoPetr/deep-to-shallow}.

 
\end{abstract}
\section{Introduction}
\label{sec:intro}

\let\svthefootnote\thefootnote
\newcommand\freefootnote[1]{%
  \let\thefootnote\relax%
  \footnotetext{#1}%
  \let\thefootnote\svthefootnote%
}

Deep Neural Networks (DNNs) have become a cornerstone of progress in computer vision, natural language processing, robotics, bioinformatics, and other fields. Their remarkable capabilities in approximating complex non-linear functions are primarily attributed to the models' depth and size~\cite{krizhevsky2012imagenet,simonyan2014very,szegedy2015going,he2016deep}. However, these same properties make training and inference computationally demanding, increasing latency, hardware costs, energy consumption, and environmental impact. This limits the deployment of DNNs in resource-constrained settings~\cite{10444008,9292253,faiz2023llmcarbon}, motivating a broad line of work on model compression and efficient inference~\cite{zhu2016trained,han2015deep,neill2020overview,astrid2018deep,howard2017mobilenets,shen2021efficient}.

\freefootnote{This article has been accepted for publication at the ITEM Workshop of the European Conference on Machine Learning and Principles and Practice of Knowledge Discovery in Databases (ECML PKDD 2026).}

\begin{figure}[t]
\centering
\includegraphics[width=1\columnwidth]{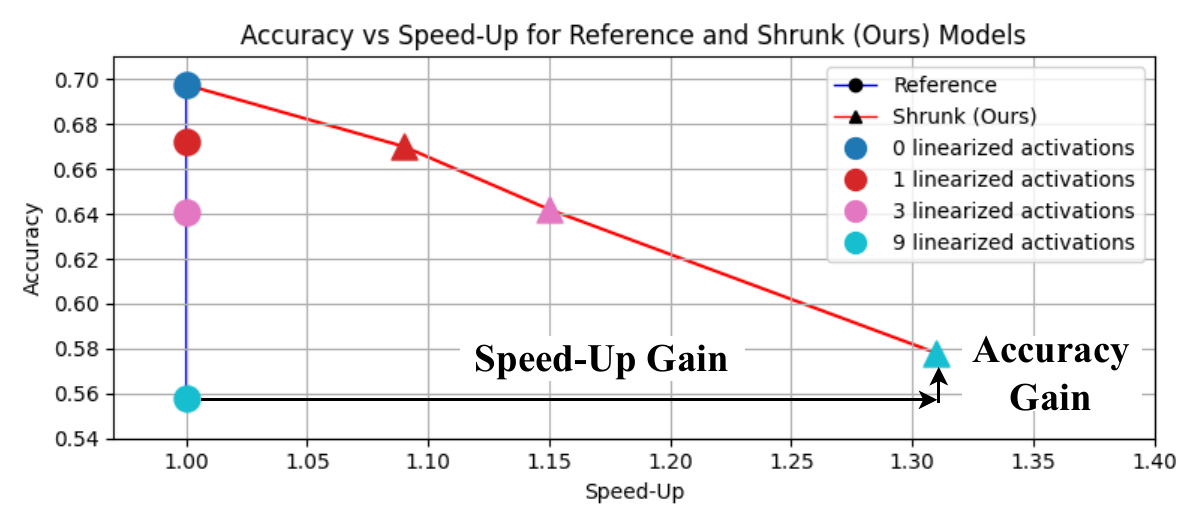}
\caption{Accuracy vs. speed-up for ResNet18 on ImageNet-1k. Unlike the Reference model, the proposed Shrunk model converts additional linearized activations into actual speed-up while preserving accuracy.}
\label{fig:teaser-accuracy-speedup-imagenet}
\end{figure}

Among compression strategies, depth compression reduces model depth by identifying redundant non-linear activations, replacing them with identity mappings, and merging the resulting consecutive linear operations~\cite{quetu2024simpler,liao2025till,liao2023can}. Yet, existing approaches face two practical limitations. First, analytical solutions for merged weights are unavailable for several common convolutional configurations, such as when the second convolution uses non-zero padding. Second, even when analytical merging is possible, the equivalent convolution may require a larger kernel, which can reduce or cancel the expected inference speed-up~\cite{pilolayerfold}. As shown in~\autoref{fig:teaser-accuracy-speedup-imagenet}, simply increasing the number of linearized activations does not improve inference time unless the corresponding blocks are effectively merged.

In this paper, we address this limitation with a loss-based layer merging strategy guided by the original linearized model, referred to as the Reference model. Instead of relying on input-independent analytical formulas, our method learns compact replacement layers that approximate both local activations and global predictions. This enables layer merging under arbitrary configurations, including padding, stride, dilation, and different kernel sizes, while avoiding the kernel growth that can limit speed-up. The resulting Shrunk model achieves increasing inference acceleration as more activations are linearized, with minimal, and in some cases no, accuracy degradation.

The contributions of this work can be summarized as follows.
\begin{itemize}
\item We introduce a generic merging strategy that applies independently of the activation linearization method, supports cases where no closed-form analytical solution is available, and avoids kernel-size growth (\autoref{tab:linearization-methods}).
\item We show that our method achieves performance close to analytical merging when such a solution exists, even with smaller and more efficient kernels (\autoref{sec:prelim-exp}).
\item We validate the approach across architectures, datasets, and hardware platforms, showing inference speed-ups on NVIDIA GeForce RTX 2080 Ti, NVIDIA Jetson Orin, and Raspberry Pi 5 with marginal or no accuracy loss (\autoref{sec:main-res}).
\end{itemize}
\section{Related Work}

Several techniques have been proposed to reduce the computational cost of deep neural networks, including quantization~\cite{zhu2016trained}, pruning~\cite{han2015deep}, low-rank approximation~\cite{astrid2018deep}, and efficient architectural blocks such as depthwise convolutions~\cite{howard2017mobilenets} or linear attention~\cite{shen2021efficient}. Among them, pruning removes redundant parameters or structures according to criteria such as magnitude, gradients, or sensitivity~\cite{lee2018snip,zhu2017prune,tartaglione2022loss}. While unstructured pruning can yield high sparsity, it often requires dedicated hardware to obtain practical speedups. Structured pruning instead removes channels, neurons, or layers, and is therefore more hardware-friendly, but may introduce a stronger accuracy–efficiency trade-off when applied aggressively~\cite{han2015learning,he2023structured,tartaglione2021serene}.

Depth compression methods aim to reduce the number of layers rather than only the number of parameters. They typically exploit activation linearization: redundant non-linearities are replaced with identity mappings, allowing the surrounding linear operations to be collapsed into a shallower model. Existing methods mainly differ in how they identify removable activations. LayerFolding~\cite{dror2021layer} and DepthShrinker~\cite{fu2022depthshrinker} introduce trainable parameters or masks to learn which activations can be linearized. EASIER~\cite{quetu2024simpler} relies on entropy to detect low-informative activations, while TLC~\cite{liao2025till} uses batch-normalization statistics to identify these activations. These approaches show that many non-linearities can be removed with limited accuracy degradation, but the resulting speed-up depends on whether the adjacent layers can actually be merged.

Layer merging is therefore the key step that turns activation linearization into practical acceleration. For consecutive linear layers, merging can be obtained by composing weights and biases. For convolutional layers, analytical merging is possible only under specific padding configurations and generally increases the effective kernel size~\cite{pilolayerfold}. In particular, when the second convolution uses non-zero padding, no input-independent closed-form solution is known, and larger merged kernels may reduce or even cancel the expected speed-up. Recent methods attempt to avoid kernel growth, but often rely on architecture-specific assumptions or jointly remove activations and convolutions~\cite{fu2022depthshrinker,kim2024layermerge}.

Our method addresses both limitations by learning merged layers under the guidance of a reference model rather than relying solely on analytical local equivalence. This enables merging configurations without closed-form solutions in a globally aware fashion. Besides this, our strategy offers greater flexibility in selecting the configuration of the replacement layer, enabling compact replacements to improve efficiency while preserving task performance.

\section{Method}
\label{sec:Method}

This section presents the proposed method for compressing deep networks by merging consecutive linear operations, \ie, linear or convolutional layers separated by previously linearized activation functions. The method supports arbitrary padding, stride, and dilation configurations, and largely preserves the original model's quality.

\subsection{Preliminaries}
\label{sec:method_preliminaries}

\begin{figure}[t]
    \centering
    \includegraphics[width=1\textwidth]{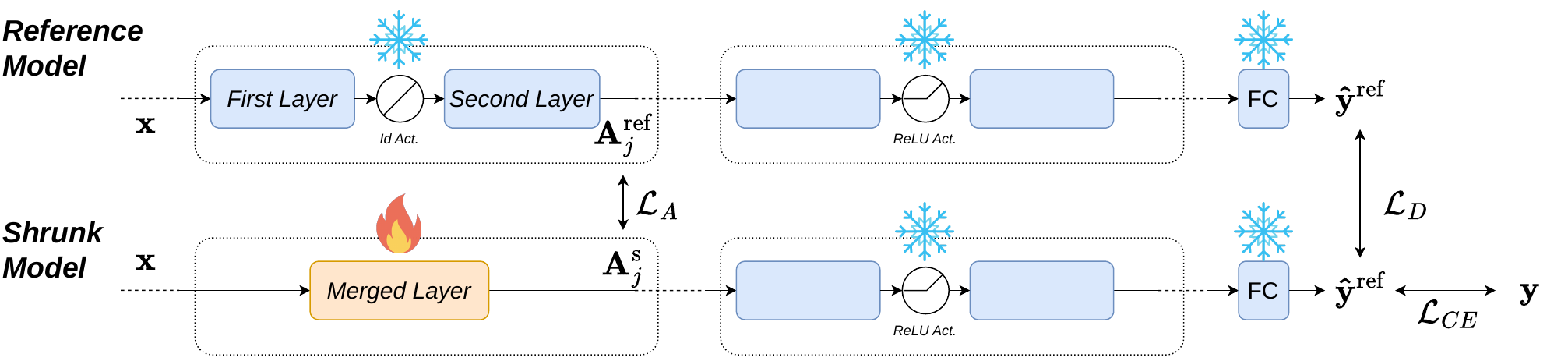}
    \caption{Illustration of the proposed method. Layers connected through linearized activations are merged and replaced by a single trainable layer. Only these layers are optimized during fine-tuning, while all others are kept frozen.}
    \label{fig:method}
\end{figure}

We consider the problem of merging consecutive linear operations separated by identity activations. Let $W_1,b_1$ and $W_2,b_2$ denote the weights and biases of two consecutive linear layers, with input/output dimensions $d^{in}_1,d^{out}_1$ and $d^{in}_2,d^{out}_2$, respectively. For convolutional layers, we denote the two operations as $\text{conv}_A$ and $\text{conv}_C$, where $\text{conv}_A$ has $c^{in}_A$ input channels, $c^{out}_A$ output channels, and kernel size $k_A$, while $\text{conv}_C$ is defined analogously. Batch normalization layers are omitted, as they can be folded into adjacent convolutions.

For fully connected layers, exact merging is obtained by composing weights and biases:
$W_{\text{merged}} = W_2 W_1$ and $b_{\text{merged}} = W_2 b_1 + b_2$. For convolutional layers, analytical merging is possible only under restricted padding configurations. When both convolutions use zero padding, or when padding is only in $\text{conv}_A$, the equivalent convolution has $c^{in}_A$ input channels, $c^{out}_C$ output channels, and kernel size $k_A+k_C-1$, which may increase computational cost. When $\text{conv}_C$ uses non-zero padding, no input-independent closed-form solution is known.

Our method handles all these cases, including padding, stride, dilation, and different kernel sizes, by learning the merged operation rather than deriving it analytically. This also allows the replacement layer size to be chosen according to the desired accuracy–efficiency trade-off.

\subsection{Shrunk Architecture Definition}
\label{sec:shapes}

The proposed method of merging consecutive linear operations is applied to models in which a certain number of activations have been linearized using an arbitrary layer collapse algorithm. For each block of model layers with activations replaced by identity functions, the weights of the merged operations are obtained through fine-tuning. 
\autoref{fig:method} demonstrates an example of the proposed approach for two generic layers. The Reference model corresponds to the original model, \ie, the pre-trained model with linearized activation functions. In parallel, a Shrunk model is constructed. The latter is identical to the Reference model except for the block(s) where layers are merged.  In the case of two fully connected layers, these layers are replaced by a single linear layer with $d_1^{\text{in}}$ input features and $d_2^{\text{out}}$ output features. In the case of convolutional layers, the block is substituted with a single convolutional layer configured to have $c_A^{\text{in}}$ input channels and $c_C^{\text{out}}$ output channels. The kernel size is set to that of the original block or max($k_A, k_C$) if multiple kernel sizes are involved, while the stride and padding are adjusted appropriately to preserve the spatial dimensions of the output and maintain architectural compatibility.

\subsection{Learning to Merge Layers}
\label{sec:fine-tuning}

Although it is possible to derive an analytical solution to match the outputs of the original and locally joined blocks, this strategy assumes that the structural constraints required for exact equivalence are satisfied (\ie, increased kernel size). In practice, however, one may want to replace a block with a more compact layer, such as a convolution with a smaller kernel size or fewer parameters, in order to improve efficiency. In these cases, it is not possible to train the new layers only locally (\ie, aligning them with the output of the replaced layers). It is also important to consider the overall impact of this new layer on the model. 
\\[.5em]
\noindent
{\bf Training Objective.} To this end, the proposed training objective does not rely solely on local alignment. As shown in \autoref{fig:method}, all layers of the Reference and Shrunk models are frozen, except for the layers of merged operations, enabling focused approximation and reducing training cost due to the limited number of trainable parameters. Batch normalization layers operate in evaluation mode, with fixed running statistics and no parameter updates. In this context, the goal is to determine the weights of the merged layers such that the Shrunk model retains comparable performance to the Reference model, taking into account both the local feature behavior and the global task-level predictions.

To achieve this, the loss function incorporates three components: a classification loss that encourages correct output predictions, a distillation loss that aligns the final logits of the Shrunk and Reference models, and an activation alignment loss that targets local alignment. More specifically, let $N$ denote the batch size and $C$ represent the number of classes. The components are defined as follows:
\begin{itemize}
\item \emph{Cross-Entropy Loss}, applied when ground-truth labels are available, between the Shrunk model predictions and ground-truth labels:
\begin{equation}
\mathcal{L}_{CE} = - \frac{1}{N} \sum_{n=1}^{N} \sum_{i=1}^{C} \mathbf{y}_i(n) \log(\hat{\mathbf{y}}_i^{s}{(n)}).
\label{eq:ce-loss}
\end{equation}

\item \emph{Distillation loss}, with temperature parameter $\tau$ between the logits predicted by the Reference model and the logits of the Shrunk model as suggested in \cite{hinton2015distilling}. The softened probabilities $\mathbf{p}_i^{\text{ref}}(n)$ and $\mathbf{p}_i^{\text{s}}(n)$ are obtained from the Reference (ref) and Shrunk (s) model by applying the softmax function with temperature $\tau$:
\begin{equation}
\mathcal{L}_{D} = \frac{\tau^2}{N} \sum_{n=1}^{N} \sum_{i=1}^{C} \mathbf{p}_i^{\text{ref}}(n) \left( \log \left( \mathbf{p}_i^{\text{ref}}(n) \right) - \log \left( \mathbf{p}_i^{\text{s}}(n) \right) \right).
\label{eq:distill-loss}
\end{equation}

\item \emph{Activation Map loss}, averaged over all merging blocks, distance-based loss between activation maps (A) after the linearized activation blocks of the Reference model and after the corresponding Shrunk model blocks:
\begin{equation}
\mathcal{L}_{A} = \frac{1}{N_{\text{blocks}}} \sum_{j=1}^{N_{\text{blocks}}} \frac{1}{N} \sum_{n=1}^{N} \left\| \mathbf{A}_j^{\text{ref}}(n) - \mathbf{A}_j^{\text{s}}(n) \right\|_2^2.
\label{eq:am-loss}
\end{equation}

\end{itemize}

Thus, the proposed loss function \eqref{eq:total-loss} is the weighted sum 
\begin{equation}
    \mathcal{L}_{Comb} = w_{CE} \mathcal{L}_{CE} + w_{D} \mathcal{L}_{D} + w_{A} \mathcal{L}_{A}.
\label{eq:total-loss}
\end{equation}
% \\[.5em]
\noindent
{\bf Training Procedure. }
The training procedure for the shrunk model is detailed in Alg.~\ref{algo:pseudocode}. 
As defined in line ~\ref{line:shrink}, the \textsc{GetShrunkModel} function is used to create our shrunk model $\mathcal{M}^{\text{s}}$. Indeed, when an identity function is found between two layers in the reference model $\mathcal{M}^{\text{ref}}$, this function replaces the entire block with a single trainable layer in the shrunk model. This layer is the only component that will be optimized during training (line~\ref{line:freeze}). The size $d$ of this new layer is chosen by the user based on the desired trade-off between efficiency and accuracy.

During training, both models are used. Indeed, for each minibatch of data, the reference model computes a forward step (line~\ref{line:fwd_ref}). Here, the model returns not only the output predictions $\mathbf{\hat{y}}^{\text{ref}}$ but also intermediate activations $\mathbf{A}^{\text{ref}}$ at predefined locations corresponding to the compressed regions. A graphical explanation is also provided in \autoref{fig:method}. The same is done for the shrunk model in line~\ref{line:fwd_shrunk}. 
The total loss, computed in line~\ref{line:loss} using the \textsc{ComputeLoss} function, consists of three components, as previously detailed in~\eqref{eq:total-loss}
Finally, in line~\ref{line:update}, the shrunk model is updated through backpropagation, but only the parameters of the new, trainable layers are modified. Notably, although this pipeline requires parallel forward passes through the Reference and Shrunk models, the overhead is confined to the training phase and does not affect the speed-up of the resulting Shrunk model during inference.

\begin{algorithm}[t]
\caption{Train Shrunk Model from linearized Reference}
\begin{algorithmic}[1]
\REQUIRE Reference model $\mathcal{M}^{\text{ref}}$, training data $(\mathbf{x}, \mathbf{y})$, layer size $d$
\STATE $\mathcal{M}^{\text{s}} \leftarrow \textsc{GetShrunkModel}(\mathcal{M}^{\text{ref}}, d)$ \label{line:shrink}
\STATE $\mathcal{M}^{\text{s}} \leftarrow \textsc{FreezeCommonLayers}(\mathcal{M}^{\text{s}},\mathcal{M}^{\text{ref}})$ \label{line:freeze}
\FOR{each minibatch $(\mathbf{x}_b, \mathbf{y}_b)$}
    \STATE $\mathbf{\hat{y}}^{\text{ref}}, \mathbf{A}^{\text{ref}} \leftarrow \textsc{Forward}(\mathcal{M}^{\text{ref}}, \mathbf{x}_b)$ 
    
    \label{line:fwd_ref}
    \STATE $\mathbf{\hat{y}}^{\text{s}}, \mathbf{A}^{\text{s}} \leftarrow 
    \textsc{Forward}(\mathcal{M}^{\text{s}}, \mathbf{x}_b)$ \label{line:fwd_shrunk}
    
    \STATE $\mathcal{L} \leftarrow \textsc{ComputeLoss}(\mathbf{\hat{y}}^{\text{ref}}, \mathbf{\hat{y}}^{\text{s}}, \mathbf{A}^{\text{ref}}, \mathbf{A}^{\text{s}}, \mathbf{y}_b)$ \label{line:loss}

    \STATE \textsc{BackpropagateAndUpdate}($\mathcal{M}^{\text{s}}, \mathcal{L}$) \label{line:update}
\ENDFOR
\end{algorithmic}
\label{algo:pseudocode}
\end{algorithm}
\section{Experiments}
\label{sec:Experiments}

In this section, we provide experimental results demonstrating the universality and effectiveness of the proposed method across diverse architectures, layer parameters, and datasets. 

\subsection{Experimental Setup}

{\bf Architectures and Training Configurations.} We conducted experiments with the following popular architectures: ResNet18~\cite{he2016deep}, ResNet50~\cite{he2016deep}, MobileNet\-v2~\cite{sandler2018mobilenetv2}, and Swin-T Transformer~\cite{liu2021swin}.
In all cases, unless otherwise stated, the models had previously been processed using the EASIER~\cite{quetu2024simpler} method for activation function linearization, though our method is applicable regardless of the specific layer-collapse algorithm used (\autoref{sec:further-analysis}). Unless stated otherwise, all models considered in the experiments use non-zero padding in the second convolutional layer, representing the general case where an analytical solution is not available. The last batch normalization layer, skip connections, and other architectural details were preserved. Padding, stride, and groups were selected to maintain spatial coherence in the activation maps. Merged convolutions were initialized either randomly or with the average weights of the merging blocks, which in some cases accelerates fine-tuning with a kernel size of $3 \times 3$. In MobileNetv2, layers in InvertedResidual blocks were merged into a single Depthwise Separable convolution when both activations were linearized. 
Loss weights and training hyperparameters were optimized only once during the CIFAR-10 experiments; therefore, further improvements in evaluation metrics may be achievable with more thorough hyperparameter tuning. The following loss weights were used during experiments: $w_{D}=1.97$, $w_{A}=0.14$, $w_{CE}=0.225$ (with $w_{CE}=0.5$ for low-accuracy models). Fine-tuning was performed using SGD~\cite{paszke2019pytorch} optimizer with momentum $0.9$, weight decay $6e-4$, and a learning rate $1e-2$.
\\[.5em]
\noindent
{\bf Datasets.} 
The models were trained and fine-tuned on the CIFAR-10~\cite{krizhevsky2009learning} and ImageNet-1k~\cite{ILSVRC15} datasets. This combination of datasets provides a variety of input image sizes, dataset sizes, styles, and representations. CIFAR-10 consists of 60k color images (32×32 pixels) from 10 classes. ImageNet-1k dataset has 1.28M training and 50k validation color images from 1000 classes, which were resized to 224x224 pixels. 
\\[.5em]
\noindent
{\bf Devices.} Inference time measurements were conducted on the following devices: NVIDIA GeForce RTX 2080 Ti with 3452 CUDA cores and 11GB GDDR6; NVIDIA Jetson Orin with 1024 CUDA cores and 8GB memory; and CPU-only Raspberry Pi 5 with 16GB RAM. Here, we consider the difference between the inference time of the reference model ($T_{\text{ref}}$) and our shrunk model ($T_{\text{shrunk}}$).

\subsection{Preliminary Experiment}
\label{sec:prelim-exp}

\begin{wraptable}{r}{0.51\textwidth}
\vspace{-33pt}
\caption{CIFAR-10 accuracy of ResNet18 Shrunk models with analytical and loss-based merging under different kernel sizes.}
\label{tab:preliminary-experiment}
\vspace{5pt}
\centering
\resizebox{0.5\textwidth}{!}{
\begin{tabular}{cccc} 
\toprule
Method &  Kernel size & Loss & Shrunk acc. \\
\midrule
Analytical solution & 5 & -- & 91.61 \\
\midrule
\multirow{3}{*}{Ours} & 5 & $\mathcal{L}_{A}$~\eqref{eq:am-loss} & 91.07 \\
& 3 & $\mathcal{L}_{A}$~\eqref{eq:am-loss} & 90.61 \\
& 3 & $\mathcal{L}_{Comb}$~\eqref{eq:total-loss} & 91.28 \\
\bottomrule
\end{tabular}
}
\vspace{-10pt}
\end{wraptable}

In this section, we present a preliminary experiment to show the effectiveness of our method. To do this, we consider a ResNet18 with 3 linearized activation functions, in which the second convolutional layer has zero padding. As explained in \autoref{sec:method_preliminaries}, this configuration allows us to adopt an analytical solution for the merged convolutional layers having a kernel size equal to $5 \times 5$~\cite{pilolayerfold}. As shown in \autoref{tab:preliminary-experiment}, our approach using the combined loss~\eqref{eq:total-loss} can achieve almost the same result as the analytical solution, even considering a smaller kernel size of $3 \times 3$.

However, if we try to only align the outputs of the replaced block~\eqref{eq:am-loss}, results drop significantly with such a small kernel size, and remain suboptimal even with a larger kernel size $(5 \times 5)$.
These results highlight the importance of looking further than local alignment when the merged layer does not faithfully replicate the structure required by the analytical formulation. Our combined loss~\eqref{eq:total-loss} provides this global perspective, allowing the Shrunk model to recover performance even when using structurally simplified or more efficient replacement layers.

\begin{wraptable}{r}{0.55\textwidth}
\vspace{-33pt}
\caption{ImageNet accuracy and inference-time reduction of Reference and Shrunk models across architectures and devices.}
\label{main-results}
\vspace{5pt}
\centering
\resizebox{0.55\textwidth}{!}{
\begin{tabular}{ccccccc}
\toprule
\multirow{2}{*}{Model} & Ref. & \# & Shrunk & \multicolumn{3}{c}{$(T_{\text{ref}}-T_{\text{shrunk}}) \uparrow$ (ms)}\\
& acc. & Merged & acc. &  ~Raspi5~  & ~Jetson~  & ~2080Ti~ \\
\midrule

    ResNet18 & 63.10 & 3 & 64.20 & +17.0 & +1.8 & +0.5   \\ 

    MobileNetv2 & 70.44 & 3 & 70.47 & +2.0 & +1.9 & +0.6   \\ 
    
    ResNet50 & 74.93 & 2 & 75.08 & +30.0 & +2.3 & +0.8   \\ 

    Swin-T & 77.54 & 2 & 77.51 & +28.0 & +1.2 & +0.3   \\ 
\bottomrule
\end{tabular}
}
\vspace{-20pt}
\end{wraptable}

\subsection{Main Results}
\label{sec:main-res}
\autoref{main-results} reports ImageNet accuracy and inference latency for several architectures whose activations were linearized with EASIER~\cite{quetu2024simpler}. Across all models, the proposed merging strategy preserves, and in some cases slightly improves, the Reference accuracy while reducing latency on all tested devices. For example, ResNet50 improves from $74.93$ to $75.08$ accuracy with a $30$~ms reduction on Raspberry Pi~5, while ResNet18 gains $+1.1$ points ($63.10 \rightarrow 64.20$) and saves $17$~ms. The largest gains are observed on resource-constrained CPU/edge devices, where depth reduction has the strongest impact on latency. The slight accuracy improvements over the Reference model are likely due to the additional fine-tuning with our combined objective: while distillation and activation alignment preserve the behavior of the linearized teacher, the cross-entropy term can correct some of its prediction errors.

\subsection{Extended Analysis}
\label{sec:further-analysis}
In this section, we provide a more in-depth investigation of our method to better understand its behavior and show its general applicability.

\vspace{.5em}

\begin{wraptable}{r}{0.5\textwidth}
\vspace{-11pt}
\caption{CIFAR-10 accuracy of Reference and Shrunk ResNet18 models with different activation linearization methods.}
\label{tab:linearization-methods}
\vspace{5pt}
\centering
\resizebox{0.5\textwidth}{!}{
\begin{tabular}{cccc} 
\toprule
Method &  \# Merged & Ref. acc. & Shrunk acc. \\
\midrule
Smallest weights & 2 & 92.20 & 92.27 \\
Smallest gradients & 2 & 92.35 & 92.40 \\
TLC & 2 & 92.63 & 92.67 \\
EASIER & 2 & 92.85 & 92.84 \\
\bottomrule
\end{tabular}
}
\vspace{-25pt}
\end{wraptable}

\noindent
{\bf Non-linearity Linearization Methods.}
Here, we conduct a comparative analysis of various activation linearization methods to demonstrate the universal applicability of the approach proposed in this work. We consider four methods that span from relatively baseline magnitude-based techniques, such as Smallest Weights / Gradients~\cite{liao2025till}, to more advanced strategies, TLC~\cite{liao2025till} and EASIER~\cite{quetu2024simpler}, which rely on the analysis of batch normalization parameters and entropy, respectively.

Experiments are conducted using the ResNet18 model on the CIFAR-10 dataset. \autoref{tab:linearization-methods} presents the test accuracies obtained after fine-tuning for each linearization method, reported for both the Reference models and the Shrunk models, in which two convolutional blocks were merged.
These results show that for all considered activation linearization strategies, the resulting model with merged layers (trained using our proposed method) achieves results in line with (or even exceeding) the Reference model.  

\vspace{.5em}

\begin{wraptable}{r}{0.5\textwidth}
\vspace{-33pt}
\centering
\caption{CIFAR-10 test accuracy of ResNet18 Shrunk models with 7 linearized activations and 4 merged blocks, evaluated across different kernel sizes used in the merged convolutional layers.}
\label{tab:kernel-size-comparison}
\vspace{5pt}
\resizebox{0.5\textwidth}{!}{
\begin{tabular}{lccccccc}
\toprule
\multirow{2}{*}{Model} & \multirow{2}{*}{Kernel} & Ref. & Shrunk & \multicolumn{3}{c}{$(T_{\text{ref}}-T_{\text{shrunk}}) \uparrow$ (ms)} \\
& & acc. & acc. &  ~Raspi5~  & ~Jetson~  & ~2080Ti~ \\
\midrule

\multirow{3}{*}{ResNet18} & 5 & 92.68 & 92.69 & -3.2 & -1.5 & -0.1 \\ 
\cmidrule(lr){2-7}
& 3    & 92.68 & 92.32 & +10.4 & +0.4  & +0.9 \\
\cmidrule(lr){2-7}
& DS3 & 92.68 & 91.79 & +16.3 & +2.5   & +0.9 \\
\bottomrule
\end{tabular}
}
\vspace{-20pt}
\end{wraptable}

\noindent
{\bf Analysis of Kernel Size Selection.}
As previously described, merging convolutional layers using an analytical solution—when such a solution exists—leads to an increase in kernel size, which can limit the achievable acceleration on real hardware. The method proposed in this work does not impose strict constraints on the choice of kernel size, as other parameters can be tuned to preserve spatial consistency.
In this section, we analyze the trade-offs between the kernel size of merged convolutional layers, the resulting test accuracy, and the achieved acceleration on hardware. Experiments are conducted using the ResNet18 model (linearized with the EASIER~\cite{quetu2024simpler} method) on the CIFAR-10 dataset.
Specifically, since ResNet18 adopts $3 \times 3$ convolutions, we consider the following kernel configurations:
\begin{itemize}
    \item $5 \times 5$: this configuration represents the size of a kernel resulting from an analytical solution.
    \item $3 \times 3$: retains the original kernel size of the merged convolutional layers.
    \item Depthwise Separable Convolution with $3 \times 3$ kernel (DS3): a more radical approach that not only merges adjacent convolutions but also replaces them with an efficient block to further improve inference speed.
\end{itemize}

\autoref{tab:kernel-size-comparison} presents the test accuracy of both Reference and Shrunk models, as well as the inference speed-up achieved by the Shrunk models on selected hardware for each kernel configuration.
Here, even when merging $4$ convolutional blocks, the model having merged convolutions with $5 \times 5$ kernels does not produce acceleration in inference.
In contrast, the $3 \times 3$ kernel and the DS3 block reduce the number of trainable parameters, which leads to a predictable trade-off between reduced test accuracy and improved inference speed.
\\[.5em]
\noindent
{\bf Kernel Size at ImageNet Scale.}
We confirm this at ImageNet scale and on more hardware, merging near-identity residual blocks of pre-trained ResNet18/34 (the inner $3{\times}3$ pair, with the intermediate activation linearized) and training only the merged $3{\times}3$ convolutions. 
The Shrunk models retain $68.05\%$ (ResNet18, $3$ blocks, ref.\ $69.76\%$) and $67.99\%$ (ResNet34, $6$ blocks, ref.\ $73.30\%$), whereas the analytical solution cannot recover accuracy without retraining (\autoref{sec:prelim-exp}).

\begin{wraptable}{r}{0.5\textwidth}
\vspace{-33pt}
\centering
\caption{ImageNet inference speed-up ($T_{\text{ref}}/T_{\text{shrunk}}$, batch~1) of ResNet Shrunk models, for the merged-layer kernel size $3{\times}3$ (ours) and the $5{\times}5$ kernel an analytical merge requires. The merged architecture is identical in both rows, isolating the effect of kernel growth.}
\label{tab:imagenet-kernel}
\vspace{2pt}
\resizebox{0.5\textwidth}{!}{
\begin{tabular}{llccc}
\toprule
Model (\#) & Kernel & ~Raspi5~ & ~J.~Nano~ & ~J.~Orin~ \\
\midrule
\multirow{2}{*}{R18 (3)} & $3{\times}3$ & $\times1.12$ & $\times1.12$ & $\times1.14$ \\
                         & $5{\times}5$ & $\times1.01$ & $\times0.79$ & $\times1.14$ \\
\midrule
\multirow{2}{*}{R34 (6)} & $3{\times}3$ & $\times1.14$ & $\times1.15$ & $\times1.17$ \\
                         & $5{\times}5$ & $\times0.96$ & $\times0.74$ & $\times1.07$ \\
\bottomrule
\end{tabular}
}
\vspace{-18pt}
\end{wraptable}

\autoref{tab:imagenet-kernel} isolates the kernel-size effect on latency: on compute-bound devices (Raspberry~Pi~5, Jetson~Nano) our $3{\times}3$ merge gives a consistent $1.12$--$1.15\times$ speed-up, while the analytical $5{\times}5$ kernel yields none and is sometimes slower than the reference; on the launch-bound Jetson~Orin the two are more similar. Avoiding kernel growth thus matters precisely on the constrained, compute-bound hardware these models target.
These findings motivate adopting $3 \times 3$ kernels as the default configuration in subsequent experiments (except for MobileNetV2, where the DS3 block remains more effective), yielding the best balance between accuracy and efficiency.

\vspace{5pt}

\begin{wrapfigure}{r}{0.6\textwidth}
\vspace{-25pt}
\centering
\includegraphics[width=0.6\textwidth]{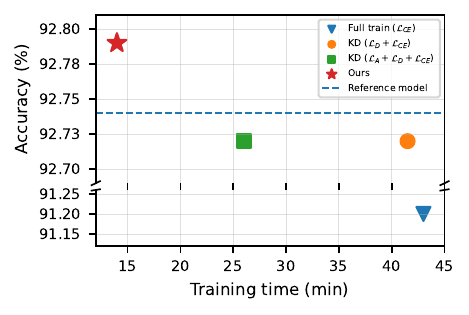}
\vspace{-25pt}
\caption{Test accuracy and convergence time of ResNet18 Shrunk models under different training strategies.}
\label{fig:kd-comparison}
\vspace{-25pt}
\end{wrapfigure}

\noindent
{\bf Knowledge Distillation Comparison.}
Given a target shrunk architecture, \autoref{fig:kd-comparison} compares our training strategy with full fine-tuning using $\mathcal{L}_{CE}$, standard KD using $\mathcal{L}_{CE}+\mathcal{L}_{D}$, and KD augmented with the activation-map loss $\mathcal{L}_{A}$. Standard KD improves over full fine-tuning, while adding $\mathcal{L}_{A}$ reduces convergence time. Our method further freezes all non-merged layers and trains only the replacement layers, reaching the best accuracy in the shortest time: $92.79\%$ in $14$~minutes, compared with $43$~minutes for full retraining and $41$~minutes for standard KD. This suggests that restricting optimization to the merged layers, while using the reference-guided loss in Eq.~\eqref{eq:total-loss}, provides faster and more stable convergence.

\vspace{5pt}

\begin{wraptable}{r}{0.4\textwidth}
\centering
\vspace{-11pt} 
\caption{Ablation of loss components for ResNet18 Shrunk models on CIFAR-10 and Tiny ImageNet.}
\label{tab:method-ablation}
\resizebox{0.3\textwidth}{!}{
\begin{tabular}{cccccc}
    \toprule
    \multirow{2}{*}{Model} & Reference &  \multirow{2}{*}{$\mathcal{L}_{\text{A}}$} & \multirow{2}{*}{$\mathcal{L}_{\text{D}}$} & \multirow{2}{*}{$\mathcal{L}_{\text{CE}}$} & Shrunk \\
    & acc. & & & & acc.\\
    \midrule

    \multirow{10}{*}{ResNet18} & \multirow{10}{*}{92.68} & -- & -- & \checkmark &  90.26   \\ 
    \cmidrule(lr){3-6}
    
    & & -- & \checkmark & -- & 92.27    \\ 
    \cmidrule(lr){3-6}
    & & \checkmark & -- & -- &  10.00  \\ 
    
    \cmidrule(lr){3-6}
    & & \checkmark & -- & \checkmark &  90.39  \\ 
    
    \cmidrule(lr){3-6}
    & & \checkmark & \checkmark & -- &  92.32  \\
    
    \cmidrule(lr){3-6}
    & & -- & \checkmark & \checkmark &  92.27    \\
    
    \cmidrule(lr){3-6}
    & & \checkmark & \checkmark & \checkmark & 92.32   \\

    \midrule

    \multirow{10}{*}{ResNet18} & \multirow{10}{*}{40.92} & -- & -- & \checkmark &  40.90   \\ 
    \cmidrule(lr){3-6}
    
    & & -- & \checkmark & -- & 41.00    \\ 
    \cmidrule(lr){3-6}
    & & \checkmark & -- & -- &  40.88  \\ 
    
    \cmidrule(lr){3-6}
    & & \checkmark & -- & \checkmark &  40.84  \\ 
    
    \cmidrule(lr){3-6}
    & & \checkmark & \checkmark & -- &  40.98  \\
    
    \cmidrule(lr){3-6}
    & & -- & \checkmark & \checkmark &  41.04    \\
    
    \cmidrule(lr){3-6}
    & & \checkmark & \checkmark & \checkmark & 41.04   \\

    \bottomrule
\end{tabular}
}

\vspace{-18pt}

% \end{table}
\end{wraptable}

\noindent
{\bf Loss Components.}

\autoref{tab:method-ablation} shows the impact of the individual loss components on the test accuracy of Shrunk models obtained by merging four blocks of the ResNet18 model using $3 \times 3$ convolution kernels in merged layers. We report two regimes: a high-accuracy ($92.68\%$ on CIFAR-10) and a low-accuracy Reference ($40.92\%$ on Tiny ImageNet).

The results show that the Activation Map loss $\mathcal{L}_A$ alone is insufficient: with four merged blocks on CIFAR-10, the Shrunk model collapses to $10.00\%$, \ie, random guessing, confirming that purely local alignment cannot recover the merged layers.

When combined with either the distillation or the cross-entropy loss, the composite objective restores accuracy to within $0.4$ points of the Reference ($92.32\%$).

When the Reference is highly accurate, the distillation term dominates and the cross-entropy term contributes little; conversely, when the Reference is weak (Tiny ImageNet), the ground-truth cross-entropy loss helps by partially compensating for the erroneous guidance of the Reference. These results corroborate the central claim of the paper: looking beyond local alignment is necessary to learn effective merged layers.

\section{Conclusion}
In this work, we introduced a layer merging method for depth-compressed networks with linearized activations, supporting configurations where analytical merging is unavailable and avoiding kernel-size growth. Experiments across architectures, datasets, and hardware platforms show that the proposed approach reduces inference latency with minimal or no accuracy degradation, and in some cases even improves over the reference model. These results indicate that the method is a practical solution for deploying compressed deep networks on resource-constrained devices.
\\[.5em]
\textbf{Acknowledgments.}
This work was supported by the French National Research Agency (ANR) in the framework of the JCJC project ``BANERA'' under Grant ANR-24-CE23-4369, by
Hi! PARIS and ANR/France 2030 program (ANR-23-IACL-0005). 


\bibliographystyle{splncs04}
\bibliography{splncs04}
%% Note that this preceding line implies that you store your BibTeX references in a file called 'mybibliography.bib'. If you instead store your references in a file with a different name, for instance 'references.bib', the preceding line should read '\bibliography{references}'. Whatever you do, DO NOT put the file name extension .bib inside the \bibliography command; this will trip up LaTeX compilers. 
%
% If you do not want to use BibTeX, you can also type up the bibliography exactly as you see fit, using the following structure:
% % Note that this number 8 reserves an amount of space (equal to the natural width of the given number) for the label of your references; if you have more than 9 references, you will want to change this number to 18. If you have more than 19 references, this number is best changed to 88. If you have more than 99 references, I salute you.




% LNCS Homepage, \url{http://www.springer.com/lncs}, last accessed 2023/10/25
% \end{thebibliography}
\end{document}